\documentclass[a4paper,fleqn]{cas-dc}

\usepackage[authoryear]{natbib}

\usepackage{float}
\usepackage{hyperref}
\usepackage{url}          
\usepackage{booktabs}       
\usepackage{amsfonts}      
\usepackage{nicefrac}      
\usepackage{microtype}      
\usepackage{graphicx}      
\graphicspath{{media/}}    
\usepackage{booktabs}
\usepackage{amsmath}
\usepackage{subfigure}
\usepackage{tikz}
\usepackage{makecell}
\usepackage{cleveref}
\usepackage{caption}
\usepackage[english]{babel}

\addto\extrasenglish{%
  \def\sectionautorefname{Section}%
  \def\figureautorefname{Figure}%
  \def\tableautorefname{Table}%
  \def\equationautorefname{Equation}%
}

\makeatletter
\let\oldappendix\appendix
\renewcommand{\appendix}{%
  \oldappendix
  \numberwithin{equation}{section}
  \numberwithin{figure}{section}
  \numberwithin{table}{section}

  \renewcommand{\sectionautorefname}{Appendix}
  \renewcommand{\equationautorefname}{Eq.}
  \renewcommand{\figureautorefname}{Fig.}
  \renewcommand{\tableautorefname}{Table}
  
  \def\equationautorefname{Eq.}
  \def\ams@eqref##1{\textup{\tagform@{##1}}} 
}
\makeatother

\def\tsc#1{\csdef{#1}{\textsc{\lowercase{#1}}\xspace}}
\tsc{WGM}
\tsc{QE}

\begin{document}
\let\WriteBookmarks\relax
\def\floatpagepagefraction{1}
\def\textpagefraction{.001}

\shorttitle{BAARC: Bayesian Adaptive Acquisition with Reconstruction-driven Classification }

\shortauthors{Adithya Ashok C.V. et~al.}  

\title [mode = title]{BAARC: Uncertainty-Driven Adaptive Partial Spectral Measurement to Full Reconstruction for Material Classification}

\author{Adithya Ashok Chalain Valapil}[orcid=0009-0004-3807-5144]

\let\printorcid\relax

\cormark[1]


\ead{adithya.ashok@uni-jena.de}

\affiliation{organization={Computer Vision Group},
            addressline={Friedrich Schiller University Jena}, 
            postcode={07743}, 
            state={Thuringia},
            country={Germany}}

\author{Aishwarya Venkataramanan}[orcid=0000-0002-6100-0034]

\author{Joachim Denzler}[orcid=0000-0002-3193-3300]

\cortext[1]{Corresponding author}


\begin{abstract}
Spectroscopic instruments with limited wavenumber coverage require accurate scan positioning or multiple measurements to complete the material analysis. This often requires prerequisite knowledge about the material under diagnosis to measure the spectral regions with distinguishing peaks. In our study, we simulate a Bayesian Adaptive Acquisition with Reconstruction-driven Classification (BAARC) framework for Raman spectral measurements, using masked training to learn spectral features that dynamically drive laser channel selection until the material is identified. We present a jointly trained two-level framework that takes progressive partial measurements of small wavenumber patches across the complete spectral range and, at each step, first attempts to reconstruct the full spectra until the model uncertainty is low. Secondly, material classification is performed at each reconstruction step until stoppage. Uncertainty quantification during reconstruction steps helps determine the measurement position of our next patch, while combining it with classification uncertainty determines measurement termination. By closing the loop between hardware acquisition and deep probabilistic modeling, the proposed system minimizes physical measurement steps while maintaining both spectral fidelity and diagnostic accuracy. Experimental evaluations demonstrate that our framework achieves high-accuracy sample identification using only a fraction of full-spectrum channels, offering a scalable solution for real-time, high-throughput spectroscopic instruments.
\end{abstract}


\maketitle

\section{Introduction}
Raman spectroscopy is a powerful analytical technique for characterizing materials through their intrinsic molecular vibrational signatures~\citep{cvr1,cvr2}. When coupled with microscopy, it enables non-destructive, label-free chemical imaging of spatially inhomogeneous specimens. However, spontaneous Raman scattering suffers from inherently weak signal intensity, requiring long integration times that severely limit its applicability to dynamic biological processes~\citep{raman,dis_raman}. Coherent anti-Stokes Raman scattering (CARS) microscopy was developed to overcome this bottleneck by enhancing signal intensity through coherent nonlinear excitation~\citep{intensity,veryinto,cars_intro}. However, CARS suffers from non-resonant background contributions that distort its spectral shape compared to spontaneous Raman~\citep{cars}. Stimulated Raman scattering (SRS) microscopy resolves this limitation by delivering background-free, quantitative spectra identical to spontaneous Raman~\citep{srs, SRS_spcov3, srs_review}. 

Despite these advancements, SRS microscopy faces a fundamental trade-off between spectral coverage, acquisition speed, and sample viability. Acquiring a full spectrum across a wide wavenumber range requires continuous laser tuning, prolonging total measurement times, increasing hardware complexity, and elevating the risk of phototoxicity in biological samples~\citep{SRS_spcov, SRS_spcov2_motiv, SRS_spcov3, mat_compl}. As a result, practical constraints often force experimentalists to acquire only partial or sparse spectral bands. Similar limitations arise across many spectroscopic modalities, such as hyperspectral imaging~\citep{appl_1}, near-infrared spectroscopy, and diffuse reflectance spectroscopy~\citep{appl_2}, where only a fraction of the available spectral information can be acquired within a given measurement budget. 

Rather than relying on uniform, exhaustive scanning, recent advances in machine learning enable a transition toward information-driven, adaptive acquisition~\citep{ml_raman, jd1, jd2}. While real-time uncertainty propagation and active sensing have been successfully applied in dynamic tracking and active vision tasks~\citep{jd_boed1, jd_boed2}, existing Raman spectroscopy methods predominantly rely on static acquisition grids. Instead of measuring every wavenumber directly, a physical system can dynamically select high-information spectral coordinates based on real-time feedback from prior measurements~\citep{appl_3}. In chemical and biological applications, reconstructing the full spectrum from partial observations is vital as it preserves physical interpretability, allowing domain experts to perform standard peak assignments and compare spectra against established reference libraries. 

To the best of our knowledge, this work introduces the first uncertainty-based, closed-loop adaptive sensing approach that uses spectral reconstruction for target classification. We call this the \textbf{Bayesian Adaptive Acquisition with Reconstruction-driven Classification (BAARC)} framework. From partial scans of a spectrum, a completely reconstructed spectrum with its prediction uncertainty is obtained. It has been shown previously that sensing in a high-uncertainty region optimally increases information gain~\citep{gain, bald}. Our autonomous measurement control follows this by measuring the next high uncertainty region at each measurement step, as shown in~\autoref{fig:method}. The reconstruction of each step is used for classification along with its uncertainty. Both uncertainties are used as stoppage conditions in our framework, automatically halting physical measurement the moment confidence in sample identification stabilizes. Bayesian uncertainty quantification offers a principled criterion for both guiding laser selection toward ambiguous spectral regions and determining when measurement can be safely halted.

We evaluate BAARC using experimentally acquired Raman benchmark datasets (RRUFF~\citep{data_rruff}, COVID-19~\citep{data_cov}, Marine Pathology~\citep{data_marine}, and Bacteria~\citep{bacteria}) to simulate hardware-in-the-loop constrained acquisition. This setting provides a controlled environment for investigating adaptive sensing strategies using various baselines (Monte Carlo Dropout (MCD)~\citep{uncert}, Neural Process (NP) \citep{np, cnp}, and Deep Gaussian Process (DGP)~\citep{vgp, dkl}) while retaining realistic spectral complexity. Our results demonstrate that the proposed framework often requires only about 50\% of the spectrum in effect to achieve the performance of a classifier trained on the full spectrum. The effectiveness of our adaptive sensing methodology is benchmarked against budget-matched random scans over the spectrum. 

The key contributions of our framework include combining a dual-level reconstruction and classification model with uncertainty-based adaptive sensing and measurement termination for spectral acquisition techniques like SRS, where spectral coverage is limited. With BAARC, we effectively reduce the amount of spectral coverage and measurement time without compromising on the material classification performance. By shifting from uniform sampling to real-time, information-driven acquisition, this approach establishes a scalable paradigm for high-throughput spectroscopic measurement.

\section{Theoretical Background} \label{sec:theory}
In this section, we discuss the theoretical motivation and major concepts behind our BAARC framework.
\subsection{\textbf{Sequential Sensing and Bayesian Experimental Design}}
Adaptive measurement of spectral channels is an intelligent data-acquisition technique in which a system dynamically updates its next measurement parameters (such as location, resolution, or integration time) based on real-time feedback from prior measurements. We formulate this adaptive spectral sampling process as a continuous-space Bayesian Optimal Experimental Design (BOED)~\citep{boed} task. BOED provides a mathematical framework to plan future measurements such that they yield the maximum possible information about an unknown system.

Let $\mathbf{x} \in \mathbb{R}^L$ represent a full continuous spectrum over $L$ wavenumbers with coordinates $\nu$, and let $\Omega_s$ denote the subset of spectral channels physically measured up to step $s$. The partially observed spectrum is represented by $\mathbf{x}_{\Omega_s}$. At each step $s$, the system seeks to select an unmeasured spectral coordinate $\nu_{s+1} \notin \Omega_s$ that maximizes the expected information gain regarding the true underlying spectral manifold $\mathbf{x}$. The next measurement patch of width $p$ is measured along $\nu_{s+1} \pm p/2$. 

In exact Bayesian inference, updating the posterior distribution over unobserved coordinates after acquiring each new wavenumber requires evaluating intractable high-dimensional integrals. To enable real-time hardware execution, our framework uses approximation methods with neural architectures~\citep{np, uncert}. For every input, these models output a predictive mean $\hat{\mathbf{x}}$ alongside point-wise reconstruction variance $\sigma^2(\nu)$.

\subsection{\textbf{Selection of Spectral Patches}}
Finding the global optimal sequence of measured wavenumber patches to maximize information gain is a combinatorial search problem. Since neighboring and chemically related spectral channels are often strongly correlated, information obtained from one region of the spectrum can reduce uncertainty in other regions. Consequently, the utility of measuring a given spectral channel depends on the set of channels that have already been observed.

According to~\citep{gain}, maximizing information gain over a submodular set function via a sequential greedy choice provides a strict performance lower bound. It is shown that a greedy acquisition strategy can achieve at least $(1-1/e)\approx63.2\%$ of the information gain obtained by the optimal offline measurement sequence. Motivated by these results, we employ predictive reconstruction variance as a computationally efficient surrogate for information gain and select the next spectral coordinate according to

\begin{equation}
    \nu_{s+1} = \arg\max_{\nu \notin \Omega_s} \sigma^2(\nu)
\end{equation}

This acquisition strategy follows the principle of uncertainty reduction commonly used in Bayesian active learning and experimental design, where measurements are preferentially acquired in regions exhibiting high predictive variance~\citep{un_var,un_var2}. Although predictive variance is not identical to information gain, it provides a practical approximation that prioritizes spectral locations where the model is expected to benefit most from additional observations.

\subsection{\textbf{Autonomous Scan Termination}}

While reconstruction variance guides spectral laser positioning, it also helps in determining when to halt physical acquisition. However, empirical evidence in our experiments shows that we primarily require a task-specific confidence evaluation. One way is to use standard predictive entropy, as it measures total uncertainty from measurement noise and epistemic uncertainty. Mathematically, the classification uncertainty can be framed as the Shannon Entropy ($H$) of the downstream class probability distribution $P(y \vert \mathbf{x}_{\Omega_s})$, where $y$ is the target class:
\begin{equation}
    H(y\vert \mathbf{x}_{\Omega_s}) = -\sum_{i} P(y_{i} \vert \mathbf{x}_{\Omega_s}) \log_{2} P(y_{i} \vert \mathbf{x}_{\Omega_s})
\end{equation}

As the system actively measures highly informative parts of the spectrum, the entropy drops. The measurement loop can terminate the instant $H(y \vert \mathbf{x}_{\Omega_s}) \leq \tau_{cfy}$, where $\tau_{cfy}$ is an acceptable error tolerance.

To explicitly isolate epistemic uncertainty (the model's lack of knowledge), we incorporate mutual information adapted from the Bayesian Active Learning by Disagreement (BALD)~\citep{bald} framework. The mutual information $I(y; \omega \vert \mathbf{x}_{\Omega_s})$ between the predicted class label $y$ and the model parameters $\omega$, given the observed channels, is expressed as:
\begin{equation}
    I(y; \omega \vert \mathbf{x}_{\Omega_s}) = \mathbb{H}[y \vert \mathbf{x}_{\Omega_s}] - \mathbb{E}_{p(\omega \vert \mathcal{D})}[\mathbb{H}[y \vert \mathbf{x}_{\Omega_s}, \omega]]
\end{equation}
where the first term is the total predictive entropy across parameter samples, and the second term represents the expected conditional entropy. Here we are subtracting the average of all individual entropies across model passes from the entropy of the averaged prediction. Thresholding this mutual information to a low value provides an information-theoretic indication that further laser probing will not significantly alter the classification decision, leading to autonomous scan termination. 

\section{Related Work}

\paragraph{\textbf{Sparse Acquisition.}}
To mitigate the slow acquisition times inherent to full-spectrum scanning, classical techniques rely on sparse sampling to measure predefined fixed patches~\citep{nmr_sparse}. Raman spectra are concentrated in sparse peaks on smooth baselines, with strong correlations between neighboring wavenumbers. Sparse-acquisition methods exploit exactly this structure to recover full spectra from far fewer measurements. Several sparse sampling methods have been applied across various spectroscopic applications~\citep{sparse_ap1, sparse_ap2, adapt_raman}. Later, dynamic sparse sampling techniques substantially accelerated chemical imaging by acquiring spectral data from only a highly informative fraction of coordinates rather than raster-scanning the entire sample field of view~\citep{cs_raman}. However, this remained a supervised approach for Raman imaging. In the domain of SRS, Pence et al.~\citep{SRS_spcov2_motiv} proposed a multi-window sparse spectral sampling approach, demonstrating that predefined, disjoint spectral windows could significantly reduce acquisition time while preserving chemical contrast. While these methods successfully compress data, they are fundamentally static and non-adaptive. Because the measurement protocol is fixed prior to the experiment, the system cannot react to the specific chemical signature being observed, potentially wasting measurement time on background noise or uninformative spectral regions.

\paragraph{\textbf{Deep Spectral Masking and Reconstruction.}}
With the rise of deep learning, masked modeling has become a powerful tool for spectral inference. Masked autoencoders~\citep{mae_base} were previously applied for reconstructing missing spectral bands~\citep{mae, spectral_mae}. They apply self-supervised learning by randomly masking portions of spectral data and training a neural network to reconstruct the missing parts. Moreover, Wu et al.~\citep{teanet} developed the TeaNet pipeline, demonstrating that applying fixed, random masks to spectra during training forces the neural network to learn highly discriminative features, ultimately improving downstream classification. However, these methods operate purely in an offline setting. They utilize static, predetermined random masks as a data augmentation or pre-training strategy on already-acquired datasets, rather than dynamically driving physical measurements.

\paragraph{\textbf{Adaptive Sensing in Spectroscopy.}}
The concept of dynamically updating instrument parameters during acquisition was first proposed by Gehm and Kinast~\citep{appl_3} to optimize spectral imaging, a concept recently operationalized by machine learning. For example, Mou et al.~\citep{rein_learn} applied Deep Reinforcement Learning (DRL) for hyperspectral band selection. However, such DRL approaches are typically task-specific, focusing strictly on classification accuracy while discarding the generative capacity to reconstruct the physical signal.  Alternatively, Bayesian active learning has been applied to analytical chemistry, such as Nabika et al.'s~\citep{sp_bayesian} framework for X-ray Photoelectron Spectroscopy (XPS), which focuses on explicit mathematical model selection for spectral curve-fitting. While Bayesian Active Learning by Disagreement (BALD)~\citep{bald} is widely used in standard machine learning to select informative data points for labeling~\citep{jd_bald}, existing active spectroscopic methods generally remain confined to rigid parameter assumptions or single-task objectives.

In contrast to the aforementioned literature, our BAARC framework bridges the gap between physical signal recovery and task-specific classification through an online, hardware-in-the-loop active sensing architecture. Unlike static protocols~\citep{cs, SRS_spcov2_motiv} or offline masking strategies~\citep{spectral_mae, teanet}, our algorithm dynamically determines which physical wavenumber to measure next based on the real-time structural ambiguity of the sample. Furthermore, avoiding the limitations of single-task adaptive methods~\citep{rein_learn, sp_bayesian}, we deploy a dual-engine system, utilizing submodular spatial variance~\citep{un_var, un_var2} to actively guide signal acquisition, while simultaneously monitoring predictive entropy and epistemic mutual information to autonomously terminate the scan once confident classification is achieved.

\section{Methods}

\begin{figure*}
    \centering
    \includegraphics[trim=1.4cm 3cm 1.5cm 2.3cm, clip=true, width=0.9\linewidth]{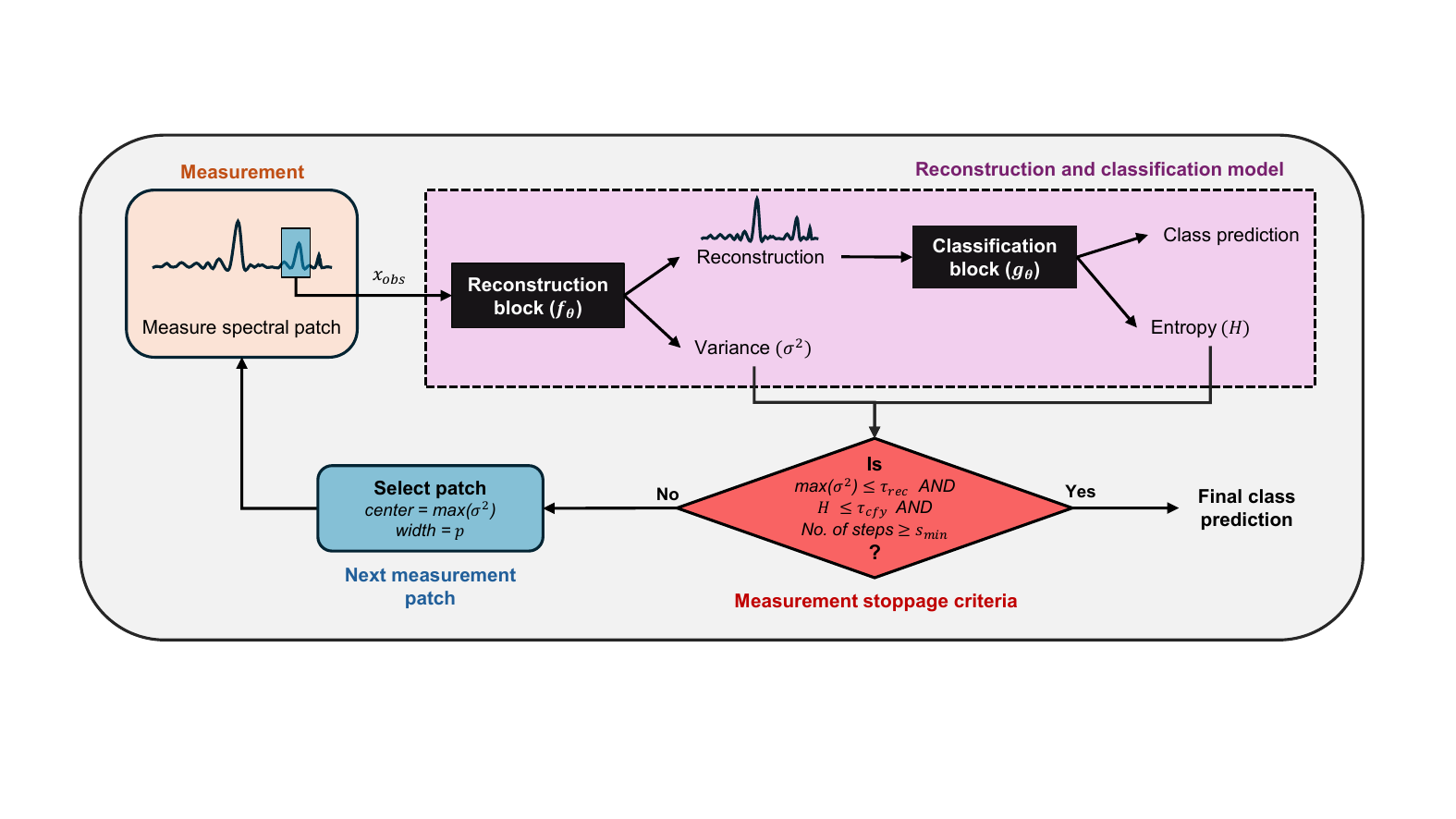}
    \caption{\textbf{BAARC measurement pipeline} shows how partial measurement of spectral patches is guided using uncertainty-driven stoppage conditions. Starting with the central patch at the initial step $s=0$, each progressive measurement step leads to full spectral reconstruction, which is used for effective classification. The uncertainty metrics are used for the stoppage condition and selecting the next measurement patch.}
    \label{fig:method}
\end{figure*}

Our methodology used in BAARC is illustrated in~\autoref{fig:method}. We input partial measurements or masked spectra to the reconstruction head in our model. Using masked training, we inpaint and complete the spectrum. This enables the classifier head to evaluate the spectrum globally rather than requiring separate models for different measurement patches based on where the measurement is done. We implement a joint training procedure for the reconstruction and classification heads. Both heads output quantified uncertainties, which are subsequently used for adaptive measurement and stopping conditions as detailed below. As introduced in~\autoref{sec:theory}, we implement uncertainty using variance, entropy, and mutual information by modeling various techniques as discussed later in~\autoref{subs:uq}. The various hyperparameter sets in our adaptive sensing procedure and is discussed in~\autoref{sec:exp}.

\subsection{Spectral Masking.} \label{para:msk}
To simulate partial spectral acquisition, we generate structured masks that remove continuous spectral regions rather than randomly masking individual and isolated wavenumber channels. For a given true spectrum $x \in \mathbb{R}^L$, we define a binary mask $m \in \{0,1\}^L$, where $m_i = 0$ is the unmasked or denotes a measured wavenumber and $m_i = 1$ denotes a masked or missing region. The observed spectrum is given by:
\begin{equation}
    \mathbf{x_{obs}} = \mathbf{x} \odot (1 - m)
\end{equation}

With each measurement step $s$, $\mathbf{x_{obs}}$ is added to the set of previously measured patches, forming $\mathbf{x}_{\Omega_s} = \{i: m_i = 0\}$. 

\subsection{Unified Reconstruction and Classification Model.}
Our model is trained on masked spectra to reconstruct a complete spectrum and subsequently classify it. This architecture is inspired by the TeaNET pipeline~\citep{teanet}. Our model has a reconstruction head $f_\theta$ and a classification head $g_\theta$. We observed that decoupling these models leads to overfitting toward a single class, when reconstructing spectra with fine class differences (see~\autoref{ap:sep_model}). Therefore, we train the models jointly. We enforce reconstruction of the measured spectra as a masked overlay of observed spectra and predicted spectra:
\begin{equation}
    \mathbf{{\hat{x}}} = \mathbf{x}_{\Omega_s} + f_\theta(\mathbf{x}_{\Omega_s}, m) \odot m
\end{equation}

A gated multi-layer perceptron (gMLP)~\citep{gmlp} architecture serves as the backbone for the reconstruction head. This is then modified based on the chosen uncertainty measures as discussed later in the next subsection. Through experimentation, we find that normal linear layers are more effective than convolutional layers for the reconstruction task and therefore constitute the architecture selection. Our classifier head uses a set of sequential linear layers. 

For $k$ classes, the classification logits are derived as:
\begin{equation}
    z = g_\theta(\mathbf{x}) ,\quad   P(y=k|\mathbf{x}) = softmax (z_k)
\end{equation}

For optimization of our model, we use a combined loss function by weighting ($\lambda$) reconstruction loss (see~\autoref{subs:uq} for specific model) and cross-entropy for classification:

\begin{equation} \label{eq:loss}
    \mathcal{L} = \lambda_{rec} \mathcal{L}_{rec} (\mathbf{x}, \mathbf{x} \odot m) + \lambda_{cfy} \mathcal{L}_{entropy} (z, y_{true})
\end{equation}

\subsection{Uncertainty quantification for active sensing.} \label{subs:uq}

We quantify uncertainty separately for the reconstruction and classification processes. For reconstruction and uncertainty quantification, we compare three architectures: Monte Carlo Dropout (MCD)~\citep{uncert}, Neural Process (NP)~\citep{np, cnp}, and Deep Gaussian Process (DGP)~\citep{vgp, dkl} as discussed below. For classification, we use only MCD, as it has low computational complexity and provides a fixed downstream architecture for easy comparison.  

\paragraph{\textbf{Monte Carlo Dropout (MCD).}} MCD randomly deactivates neurons during training and testing, enforcing a distribution over neural network weights. By adding them as dropout layers, the model stops memorizing the data by creating a different random path through the network each time. We perform $T$ stochastic forward passes by keeping the dropout layers active during inference, allowing variability in the results. The mean of these predictions serves as the reconstructed spectrum, and the epistemic uncertainty is captured by the predictive variance ($\sigma_i^2$) at each wavenumber $i$ across the $T$ passes. For classification, we compute the Shannon Entropy ($H$)~\citep{shannon, entropy} of the averaged predictive probability distribution. As discussed in~\autoref{sec:theory}, Mutual Information (MI) can additionally be utilized for the same, however, our experiments indicated that both measures follow a similar trend in this domain of work (see~\autoref{ap:mi}). Therefore, we only use it optionally. 

For MCD the reconstruction loss in~\autoref{eq:loss} is mean square error (MSE):
\begin{equation}
    \mathcal{L}_{rec} =\mathcal{L}_{MCD} = \frac{1}{L}\sum _{i=1}^{L}(\mathbf{x}_i - \mathbf{\hat{x}}_i)^{2}
\end{equation}

\paragraph{\textbf{Deep Gaussian Process (DGP).}} 
A standard Gaussian Process (GP) defines a non-parametric prior over continuous spectra. However, calculating exact GP posterior uncertainty over a full spectrum of $L$ wavenumber channels requires inverting an $L \times L$ covariance matrix, which is computationally expensive ($\mathcal{O}(L^3)$). To overcome this bottleneck, our DGP framework integrates a mask-aware gMLP encoder with a Variational Gaussian Process (VGP)~\citep{vgp, dkl}. The gMLP encoder acts as a feature compressor, mapping high-dimensional sparse observations $\mathbf{x}_{obs}$ into a low-dimensional latent embedding. The VGP then operates over this latent space using a Mat\'ern kernel, which models physical correlations between wavenumbers. To accelerate inference, the VGP uses a small, learnable set of $M$ landmark pseudo-measurements ($M \ll L$) called inducing points ($u$). By evaluating covariance exclusively across these $M$ landmark points, the matrix inversion cost drops to $\mathcal{O}(LM^2)$. 

The DGP parameters and inducing locations are trained jointly with a loss term in~\autoref{eq:loss} by maximizing the Variational Evidence Lower Bound (ELBO):

\begin{equation}
\begin{split}
    \mathcal{L}_{rec} =\mathcal{L}_{DGP} = \mathbb{E}_{q(\mathbf{f})}\left[\log p(\mathbf{x_{obs}} \mid \mathbf{f})\right] - \\\text{KL}\Big(q(u) \parallel p(u)\Big)
\end{split}    
\end{equation}

where $\mathbf{f}$ is the curve function for every spectrum and $q(\mathbf{f})$ used in the log-likelihood is the variational distribution over all of these function values. The $p(u)$ and $q(u)$ are the prior and estimates over inducing variables, and $\text{KL}(\cdot)$ denotes the Kullback-Leibler divergence. Despite inducing point optimizations, drawing posterior samples across iterative sampling steps still introduces noticeable computational overhead compared to purely neural architectures.

\paragraph{\textbf{Neural Process (NP).}} 
To preserve the non-parametric flexibility of Gaussian Processes without incurring matrix inversion latency, we implement a Latent Neural Process framework~\citep{np, cnp}. NPs include Bayesian inference directly into a deep neural network, enabling single forward-pass uncertainty estimation. In NP terminology, the input spectrum is split into two subsets: the set of physically measured wavenumber coordinates and their corresponding intensities are context points ($\mathcal{C}$), and the full set of spectral coordinates that the model must reconstruct are target points ($\mathcal{T}$). A gMLP context encoder aggregates the context points $\mathcal{C}$ into a global latent representation $z \sim q(z \mid \mathcal{C})$. During training, a parallel target encoder observes the complete spectrum $\mathcal{T}$ to produce a target latent posterior $q(z \mid \mathcal{T})$. A lightweight neural decoder then takes the latent sample $z$ alongside target wavenumber coordinates to output both the mean spectral intensity $\hat{x}$ and a predictive variance $\sigma_i^2$ in a single deterministic pass. 

The NP is trained with a loss term in~\autoref{eq:loss} by minimizing the negative log-likelihood of target reconstruction combined with a latent KL-divergence penalty:

\begin{equation}
\begin{split}
 \mathcal{L}_{rec} =\mathcal{L}_{NP} = -\mathbb{E}_{q(z \mid \mathcal{T})}\left[\log p(\mathbf{x} \mid z, \mathcal{T})\right] + \\ \text{KL}\Big(q(z \mid \mathcal{T}) \parallel q(z \mid \mathcal{C})\Big)
 \end{split}
\end{equation}

By forcing the context latent distribution $q(z \mid \mathcal{C})$ to match the target distribution $q(z \mid \mathcal{T})$, the NP learns to infer full-spectrum details and spatially resolved variance directly from sparse physical measurements without requiring iterative posterior sampling during inference.

\subsection{\textbf{Simulated adaptive sensing strategy.}} \label{para:active}

Following the acquisition of an initial central spectral patch (simulating a narrow-band SRS measurement), the model reconstructs and classifies the spectrum. Acquisition continues recursively until both reconstruction and classification uncertainties fall below predefined thresholds ($\tau_{rec}$ and $\tau_{cfy}$ respectively). 

High-variance regions indicate high reconstruction uncertainty and are prioritized for physical measurement. High entropy indicates that the currently measured patches lack sufficient discriminative features to confidently identify the class. To prevent premature termination due to model overconfidence, we enforce a minimum number of measurement steps ($s_{min}$). 

To start sensing, we select the central patch in the spectrum as our first patch at $s=0$. At each acquisition step $s$, the greedy optimization logic proceeds as follows:
\begin{enumerate}
    \item Identify the unmeasured coordinate with the maximum reconstruction variance: $\arg\max_{i \in \{i \mid m_i = 1\}} \sigma_i^{2}$.
    \item Update the mask $m$ by integrating the newly acquired spectral patch of width $p$ centered at the identified target $\nu_{s+1}$. 
    \item Perform a forward pass to update the reconstructed spectrum and classification probabilities.
    \item Repeat steps 1-3 until the maximum reconstruction variance satisfies $\max(\sigma_i^2) \leq \tau_{rec}$, the classification entropy satisfies $H \leq \tau_{cfy}$, and the step count satisfies $s \geq s_{min}$.
\end{enumerate}

Once these criteria are met, the physical scan autonomously terminates, and the final sample class is predicted. The sensing procedure can be clearly observed from the measurement steps illustrated in~\autoref{ap:rec}. The specific threshold values are empirically calibrated depending on the underlying dataset and model configuration (detailed in~\autoref{sec:exp}).

\section{Evaluation Setup} \label{sec:exp}
In this section, we present our evaluation of the models across various datasets. This also includes certain hyperparameter selection for the models' various thresholding values across our pipeline.

\begin{figure*}[t]
\centering  
\subfigure[]{\label{fig:var}\includegraphics[height=6cm,width=8cm]{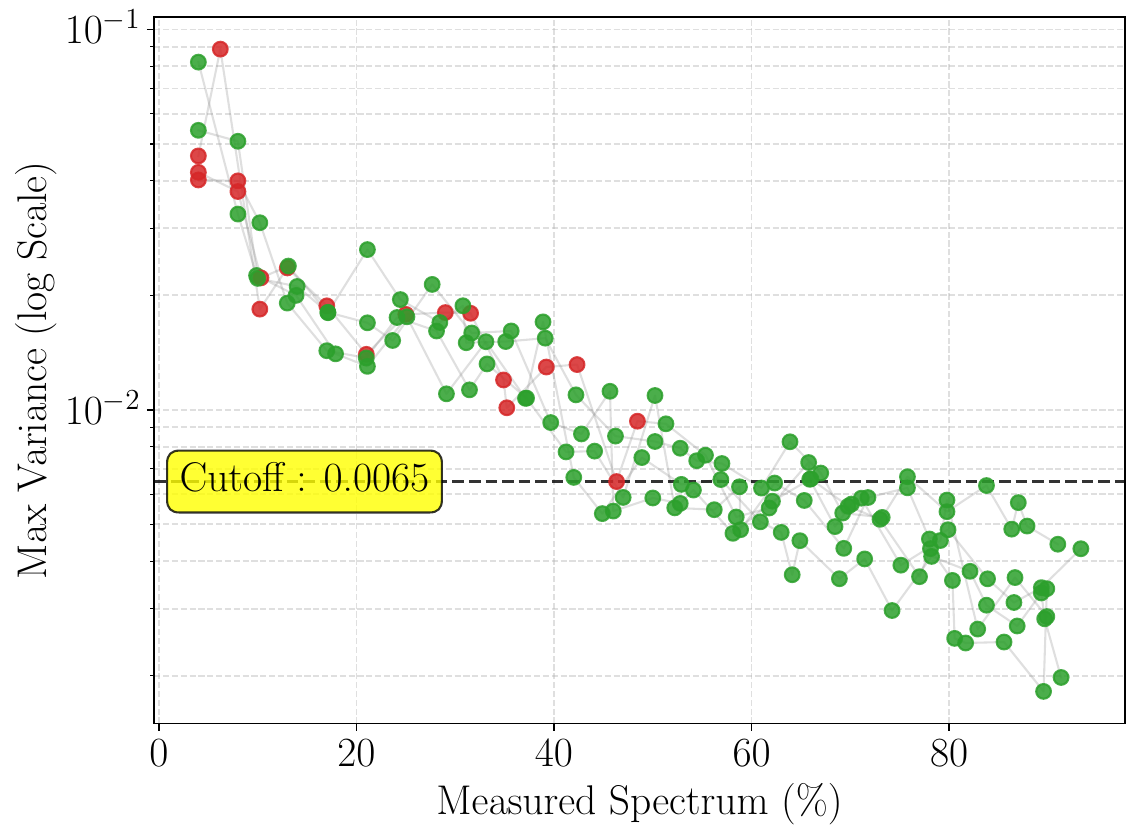}}
\hfill
\subfigure[]{\label{fig:ent}\includegraphics[height=6cm, width=8cm]{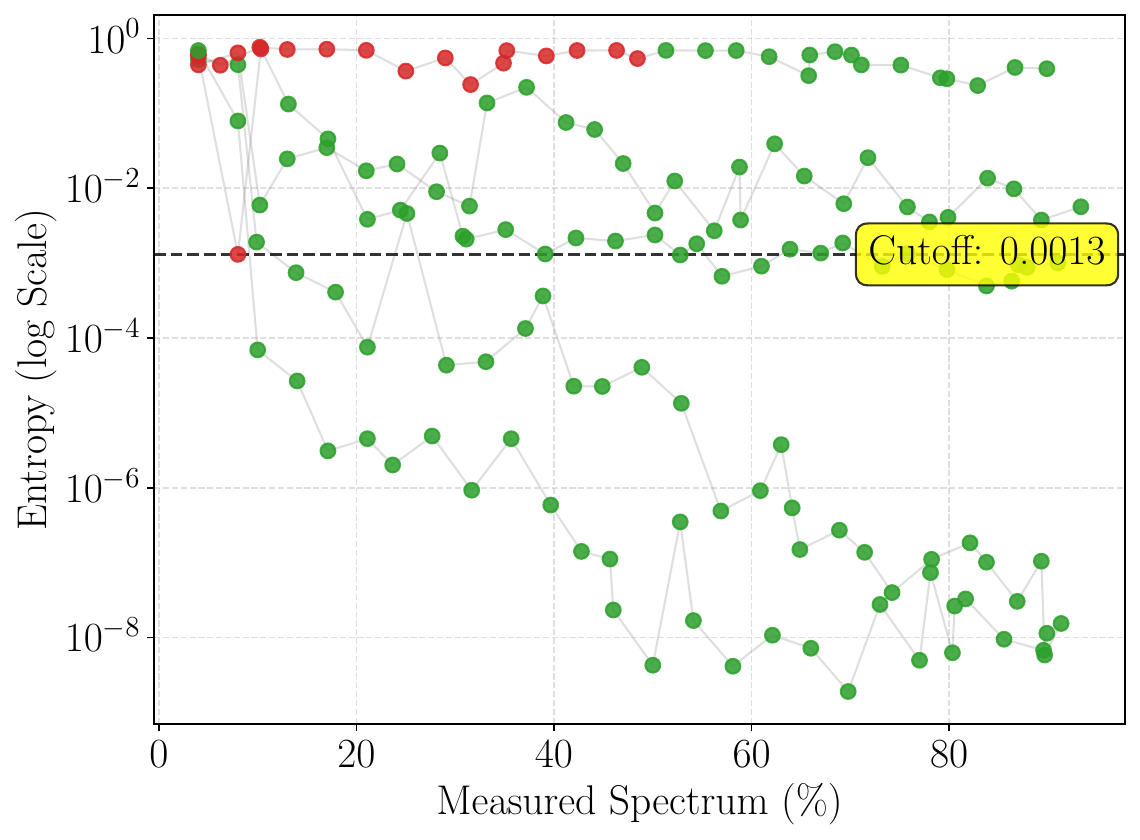}}
\caption{\textbf{Threshold selection} for stoppage condition in adaptive sensing: prediction of 5 COVID-19 spectra are plotted with (a) Maximum variance ($\sigma_{max}$) and (b) Entropy ($H$) against percentage of spectrum measured over the steps. At each step, more spectral patches are added, progressively increasing the measured percentage. We see that the falsely predicted spectral classes (red) are corrected to the true class (green) with aggregation of steps. The last false classification is used to select the cutoff.}
\label{fig:threshold}
\end{figure*}

\paragraph{\textbf{Datasets.}}
We use four publicly available Raman benchmark datasets RRUFF~\citep{data_rruff}, COVID-19~\citep{data_cov}, Marine Pathology~\citep{data_marine}, and Bacteria~\citep{bacteria} for this study. They vary widely in the number of classes, data size, spectral range, and signal-to-noise ratio (SNR). A dataset summary is given in~\autoref{app:data}.

\paragraph{\textbf{Preprocessing and data preparation.}}
Raw Raman spectra are corrected for baseline drift using asymmetric least-squares (ALS)~\citep{als}. Training samples are generated with the random masking strategy described in~\autoref{para:msk}: for each spectrum, we uniformly sample a mask covering 30-70\% of the wavenumber range, so that the model learns to reconstruct chemical peaks in the unmeasured region and, implicitly, the spectral manifold that links one region of the spectrum to another.

\paragraph{\textbf{Training.}}
The unified reconstruction-classification model is trained until both classification accuracy and reconstruction loss decrease, with hyperparameters listed in~\autoref{ap:values}. For the Bacteria dataset, we additionally fine-tune the classification head for 250 epochs, following the protocol of \citep{ramnet, pace}. The loss weights in~\autoref{eq:loss} are set to $\lambda_{\text{rec}}=2$ and $\lambda_{\text{cfy}}=1$. Most hyperparameters (learning rate, batch size, kernel width, etc.) are shared across all dataset-model pairs. The few dataset-specific values are also listed in~\autoref{ap:values}.

\paragraph{\textbf{Sensing protocol and uncertainty quantification.}}
In SRS with multi-channel detectors, a region in the spectrum is detected, capturing the local shape. We simulate these wavenumber scans over a $p=4\%$ spectral range or patch width in our experiments. At every step, the model performs $T=20$ stochastic forward passes to estimate the MCD-based predictive uncertainty.

\paragraph{\textbf{Stopping criteria.}}
Sensing halts when the reconstruction uncertainty $\max(\sigma_i^2) \leq \tau_{rec}$, the classification entropy $H \leq \tau_{cfy}$, and the step count $s \geq s_{min}$ are satisfied. 

\paragraph{\textbf{Threshold Selection.}}
\autoref{fig:threshold} illustrates the stopping-criterion selection for the COVID-19 / MCD pair, with an example showing the variation of five validation spectral sets. At each 4\% measurement step, the model produces a prediction. We mark it green (correct) or red (incorrect). The last red prediction along the variance axis occurs at $\max(\sigma_i^2)\approx 0.0065$ (\autoref{fig:var}), so we set $\tau_{\text{rec}}=0.003$, comfortably below this value to avoid false stops. For classification entropy (\autoref{fig:ent}), the minimum achievable value is $\approx 0.33$. We set $\tau_{\text{cfy}}=0.1$, which is below this floor and ensures that the entropy criterion is not the binding constraint before $s_{\min}$. In practice, we select a larger validation set ($\sim{10-20}$ spectra) for each dataset-model pair and perform the same analysis to find the thresholds. The actual values are in~\autoref{ap:values}.

Using our model pipeline, we compare three modes of measurement simulations: (i) \textbf{Full:} We use 100\% or complete spectra for classification, (ii) \textbf{Adaptive:} We follow our active learning strategy as mentioned above in this section and (iii) \textbf{Random:} We measure the next patch randomly, checking if the percentage of measured spectra is more than the average percentage of measured spectral length in the adaptive scenario. After this stoppage criterion is met, classification is done on the randomly sampled data. In this study, we compare our results with those of the respective full-spectrum mode, assuming the model is optimized for classification as well as reconstruction. The reconstructions over each measurement step are illustrated in~\autoref{ap:rec}.

\begin{figure*}[h]
    \centering
    \includegraphics[width=1\linewidth]{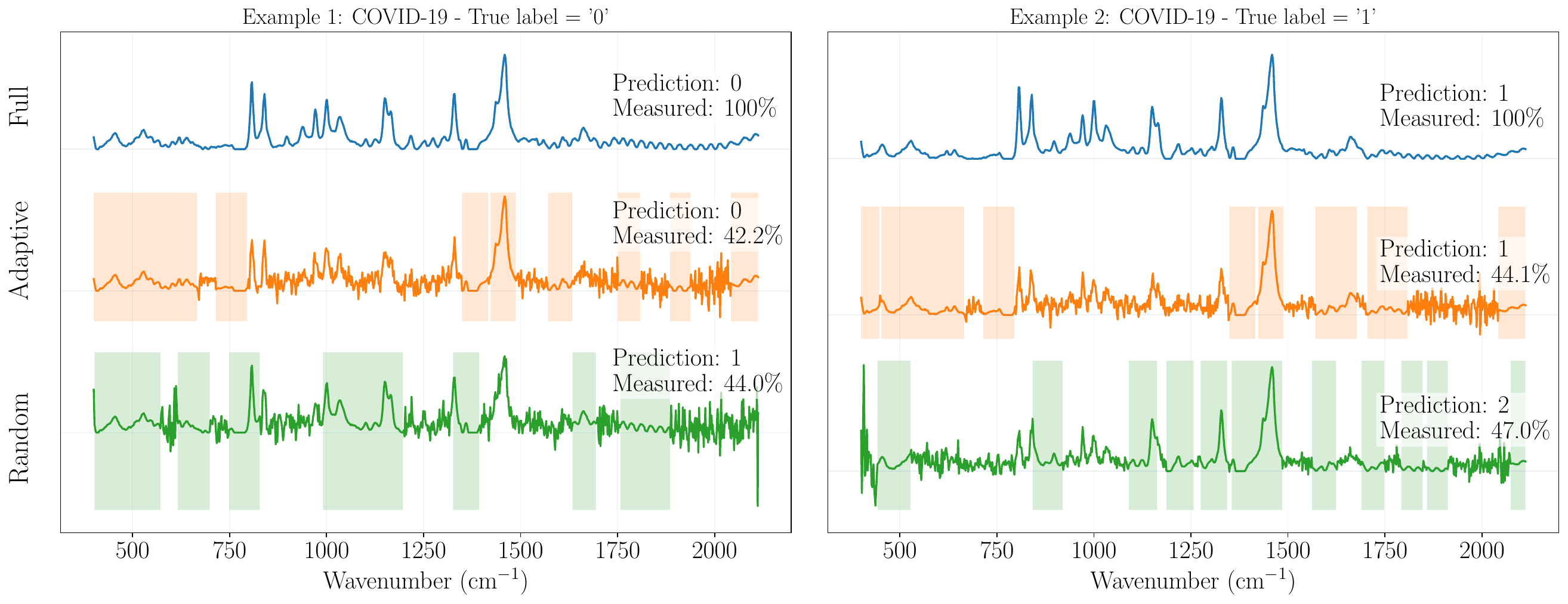}
    \caption{\textbf{Comparison of full, adaptive, and random scans} for two selected examples from the COVID-19 dataset where we have three class labels (0, 1, and 2). Examples 1 and 2 belong to true labels 1 and 0, respectively. The measured regions for adaptive and random scans are shaded. The adaptive scan makes the correct prediction with fewer than 50\% of the measured spectra, whereas the random scan fails in both examples.}
    \label{fig:cov}
\end{figure*}

\section{Results}

\begin{figure*}[thp]
    \centering
    \includegraphics[width=0.9\linewidth]{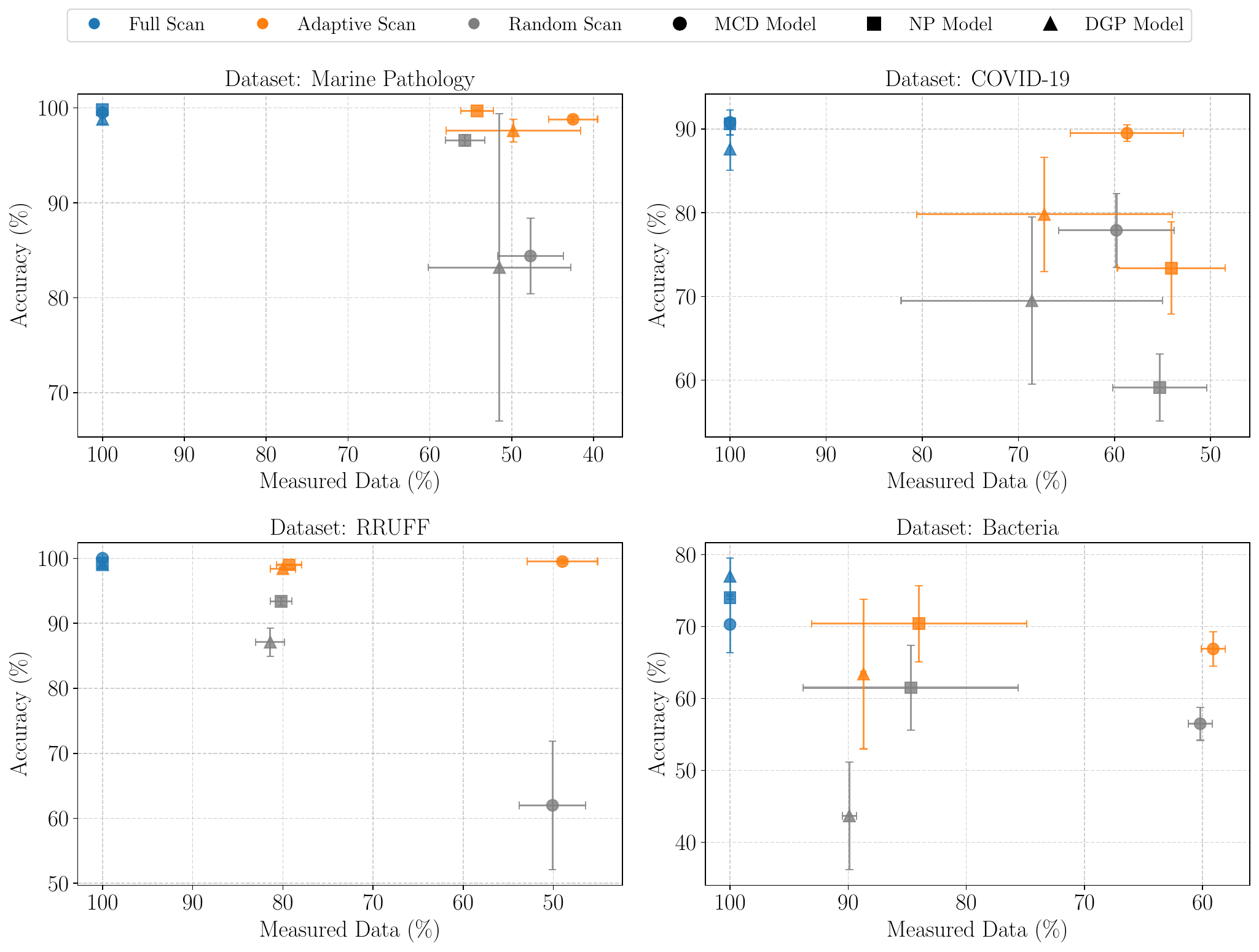}
    \caption{Our dual objectives: accuracy and measured data(\%) are plotted for each dataset to better understand the optimal scenario. The colors blue, orange, and grey represent full, adaptive, and random scan modes, respectively. The shapes represent the models MCD, NP, and DGP. The plot also shows deviations along both axes, as reported in~\autoref{fig:res}. The high-performing models appear towards the top-right corner with high accuracy and low data measured. For example, adaptive scan using MCD (orange circle) performs best across all datasets, as it is plotted towards the highest accuracy and lowest measured data. }
    \label{fig:res}
\end{figure*}

The adaptive measurement paradigm is compared with the classification of full-spectrum and randomly measured spectra as shown in~\autoref{fig:res}. In~\autoref{tab:res}, we detail the performances of MCD, NP, and DGP models for all four datasets. 

\renewcommand{\arraystretch}{1.1}
\begin{table*}[thb]
\centering
\resizebox{\textwidth}{!}{%
\begin{tabular}{ll|rrr|rrr|rrr|rrr} 
\toprule
 &Dataset:&  \multicolumn{3}{c|}{\textbf{Marine pathology}} & \multicolumn{3}{c|}{\textbf{COVID-19 }} & \multicolumn{3}{c|}{\textbf{RRUFF} } & \multicolumn{3}{c}{\textbf{Bacteria }}\\
 & & \multicolumn{3}{c|}{(Total=570)}& \multicolumn{3}{c|}{(Total=93)}& \multicolumn{3}{c|}{(Total=3554)}& \multicolumn{3}{c}{(Total=3000)}\\  
\midrule 
 Model&Measurement :& Full & Adaptive & Random  & Full & Adaptive & Random & Full & Adaptive & Random  & Full & Adaptive &Random  \\
\midrule 
&Misclassifications   & 3.0\scriptsize{$\pm$1.1}& \textbf{7.6\scriptsize{$\pm$2.2}}& 90.0\scriptsize{$\pm$22.7}& 9.1\scriptsize{$\pm$1.4}& \textbf{9.8\scriptsize{$\pm$1.0}}& 19.7\scriptsize{$\pm$4.1}& 0.0\scriptsize{$\pm$0.0}& \textbf{15.0\scriptsize{$\pm$17.7}}& 1171.3\scriptsize{$\pm$352.8}& 893.5\scriptsize{$\pm$118.3}& \textbf{997.0\scriptsize{$\pm$73.9}}&1303.2\scriptsize{$\pm$70.4}\\
\textbf{MCD} &Measured (\%)       & 100.0\scriptsize{$\pm$0.0}& \textbf{42.9\scriptsize{$\pm$4.3}}& 47.9\scriptsize{$\pm$2.7}& 100.0\scriptsize{$\pm$0.0}& \textbf{58.3\scriptsize{$\pm$6.0}}& 59.0\scriptsize{$\pm$5.7}& 100.0\scriptsize{$\pm$0.0}& \textbf{49.2\scriptsize{$\pm$3.9}}& 50.6\scriptsize{$\pm$3.6}& 100.0\scriptsize{$\pm$0.0}& \textbf{59.1\scriptsize{$\pm$1.0}}&60.2\scriptsize{$\pm$1.0}\\
&Accuracy (\%)            & 99.4\scriptsize{$\pm$0.2}& \textbf{98.8\scriptsize{$\pm$0.3}}& 84.4\scriptsize{$\pm$3.9}& 90.1\scriptsize{$\pm$1.5}& \textbf{89.4\scriptsize{$\pm$1.1}}& 78.8\scriptsize{$\pm$4.4}& 100.0\scriptsize{$\pm$0.0}& \textbf{99.5\scriptsize{$\pm$0.5}}& 67.0\scriptsize{$\pm$9.9}& 70.2\scriptsize{$\pm$3.9}& \textbf{66.7\scriptsize{$\pm$2.4}}&56.5\scriptsize{$\pm$2.3}\\
\midrule

 & Misclassifications& 5.8\scriptsize{$\pm$3.6}& 13.3\scriptsize{$\pm$6.4}& 93.6\scriptsize{$\pm$91.8}& 13.8\scriptsize{$\pm$4.3}& 22.5\scriptsize{$\pm$3.3}& 33.8\scriptsize{$\pm$5.2}& 33\scriptsize{$\pm$0.8}& 55.5\scriptsize{$\pm$11.8}&455.6\scriptsize{$\pm$74.5}& 671.5\scriptsize{$\pm$75.4}& 1099\scriptsize{$\pm$313.9}&1686.2\scriptsize{$\pm$225.8}\\
\textbf{DGP}& Measured (\%)       & 100.0\scriptsize{$\pm$0.0}& 49.5\scriptsize{$\pm$8.7}& 51.5\scriptsize{$\pm$8.7}& 100.0\scriptsize{$\pm$0.0}& 68.5\scriptsize{$\pm$5.5}& 70.1\scriptsize{$\pm$5.6}& 100.0\scriptsize{$\pm$0.0}& 80.3\scriptsize{$\pm$1.4}&81.8\scriptsize{$\pm$1.6}& 100.0\scriptsize{$\pm$0.0}& 88.7\scriptsize{$\pm$0.2}&89.9\scriptsize{$\pm$0.6}\\
 & Accuracy (\%)            & 98.9\scriptsize{$\pm$0.6}& 97.6\scriptsize{$\pm$1.2}& 83.5\scriptsize{$\pm$16.1}& 85.1\scriptsize{$\pm$4.7}& 75.8\scriptsize{$\pm$3.6}& 63.6\scriptsize{$\pm$5.6}& 99.0\scriptsize{$\pm$0.2}& 98.4\scriptsize{$\pm$0.2}&87.1\scriptsize{$\pm$2.2}& 77.6\scriptsize{$\pm$2.5}& 63.3\scriptsize{$\pm$10.4}&43.7\scriptsize{$\pm$7.5}\\ 
\midrule

  &Misclassifications& 1.0\scriptsize{$\pm$0.0}& 4.0\scriptsize{$\pm$2.4}& 16.0\scriptsize{$\pm$4.3}& 11.3\scriptsize{$\pm$2.7}& 19.1\scriptsize{$\pm$6.6}& 28.6\scriptsize{$\pm$9.7}& 24.1\scriptsize{$\pm$9.3}& 30.5\scriptsize{$\pm$4.7}& 209.3\scriptsize{$\pm$25.4}& 775.3\scriptsize{$\pm$7.0}& 885.6\scriptsize{$\pm$161.5}&1141.3\scriptsize{$\pm$177.0}\\
\textbf{ NP}&Measured (\%)       & 100.0\scriptsize{$\pm$0.0}& 55.3\scriptsize{$\pm$1.8}& 56.3\scriptsize{$\pm$1.8}& 100.0\scriptsize{$\pm$0.0}& 67.4\scriptsize{$\pm$13.8}& 68.5\scriptsize{$\pm$13.6}& 100.0\scriptsize{$\pm$0.0}& 79.3\scriptsize{$\pm$1.9}& 80.4\scriptsize{$\pm$1.2}& 100.0\scriptsize{$\pm$0.0}& 84.0\scriptsize{$\pm$9.3}&84.7\scriptsize{$\pm$9.1}\\
 &Accuracy (\%)            & 99.8\scriptsize{$\pm$0.0}& 99.2\scriptsize{$\pm$0.4}& 97.1\scriptsize{$\pm$0.7}& 87.8\scriptsize{$\pm$2.9}& 79.3\scriptsize{$\pm$7.1}& 69.1\scriptsize{$\pm$10.4}& 99.3\scriptsize{$\pm$0.3}& 99.1\scriptsize{$\pm$0.1}& 94.1\scriptsize{$\pm$0.7}& 74.1\scriptsize{$\pm$0.2}& 70.4\scriptsize{$\pm$5.3}&61.9\scriptsize{$\pm$5.9}\\
\bottomrule 
\end{tabular}
} 
\caption{Comparison of classification performance (6-fold CV) of adaptive sensing pipeline with full and random scans for four datasets: Marine pathology~\citep{data_marine}, COVID-19~\citep{data_cov}, RRUFF~\citep{data_rruff}, and Bacteria~\citep{bacteria} with models MC Dropout, Neural Process, and Deep Gaussian Process. As metrics, we report the misclassifications out of the total, the percentage of data measured, and the classification accuracy for three runs with each dataset-model pair. Considering our dual objective of least measured with best accuracy for adaptive close to full scan, we bold the best performing model values.}
\label{tab:res}
\end{table*}

We first examine two example measurements from the COVID-19 dataset using the MCD model, as shown in~\autoref{fig:cov}. We can see that adaptive measurement predicts the correct label as the full prediction. In the two examples we show, the random prediction uses the same amount of data as the adaptive, but predicts the wrong class. The scanning mode also affects the reconstruction, as is evident qualitatively in the figures. However, when quantitatively analyzed, the adaptive and random scan had similar reconstruction errors and correlation values (see~\autoref{ap:rec_app}). This is because fine differences in peaks are rarely captured by quantitative metrics, but these details become evident for fine-grained classification. Therefore, even though ours is a dual-objective model, we primarily focus on the downstream classification task rather than on optimal reconstruction. However, the quantitative analysis shows that NP achieves the most accurate reconstruction, with the highest correlation and lowest loss. This shows the possibility of an intermediate spectrum before classification, which the user can refer to during analysis.

For the \textbf{Marine pathology dataset}, we see that MCD requires only 43\% of the full spectral scan to achieve accuracy over 98\%. However, NP measures 10\% more data and achieves the highest accuracy with the lowest performance deviation, while DGP measures 7\% more data than MCD. From misclassifications in~\autoref{tab:res}, it is clear from all three models that the random scan performs the worst, especially for MCD and DGP. From~\autoref{fig:res}, we see that the least amount of data is measured for this dataset, yet it achieves high adaptive performance across all models.

For the \textbf{COVID-19 dataset}, we see that our adaptive measurement technique using MCD achieves nearly the same accuracy as full-spectrum classification while measuring 58\% of the spectrum. This shows the potential to achieve optimal classification by identifying relevant regions during spectral measurement. In our analysis, we found that the same spectra are misclassified by both the full and adaptive scanning techniques on this dataset. MCD is also at least 10\% better in accuracy than NP and DGP. However, adaptive measurement is much better than random measurement overall.

The \textbf{RRUFF} dataset is the most divergent, with 1310 classes. Here, we see that our MCD model accurately classifies all points (over 6-fold CV) when the full spectrum is provided. In adaptive measurement, misclassifications occurred at a rate of only 15, with half of the spectra measured on average, while 1171 spectra were misclassified in random measurement. A difference of less than 1\% between the full-adaptive case and about 25\% between the adaptive-random case (except NP) again justifies the strength of adaptive sensing and uncertainty-aware spectral inpainting, along with the classifier. It should be noted that NP and DGP measured ~80\% of the data to achieve this performance, while MCD achieved the highest performance using only 50\% of the data.

The \textbf{Bacteria dataset} is the largest, with over 60000 spectra, and has 30 classes. All models are instructed to be fine-tuned with the given fine-tuning batch~\citep{ramnet, pace}. The weakest performance is observed here. For the adaptive model, the classification accuracy of MCD is 66\% with measuring 60\% of the data. Note that NP performs better at 70\% accuracy but requires 84\% of the data, and the performance difference between adaptive and full scan for MCD and NP is only 4\%.

\section{Discussion}
Overall, all three adaptive models across all four datasets demonstrate that adaptive sensing is far superior to random scanning and often approaches the performance of a full scan with significantly less data. The adaptive scan performance complements the full-scan classification accuracy, supporting our major assumption that adaptive sensing is well-suited only to scenarios where full-spectral classification is possible. Apart from their lowest performance, random scans are often observed to have a high standard deviation. 

From~\autoref{fig:res}, we can see that MCD performs best across all datasets, appearing at the top-right corner of the plot. Moreover, when simplicity of implementation and low computation time are also taken into account, MCD clearly performs best among the baselines, followed by NP and DGP. It is also clear that MCD required significantly less data to achieve peak performance than the others and had comparatively coarse or relaxed adaptive hyperparameters. We also shortly explore the credibility of our intermediate reconstructed spectra from the model. This is relevant to a spectroscopist as it gives insight for further analysis rather than just a classification output. Here, we found that NP performed best.     

One major limitation in our pipeline is that, during adaptive sensing, the model often does not measure the whole spectrum, leading to potential false predictions. However, in most cases, we see that false predictions occur even when the full-spectrum classifier also fails, despite having the full spectrum. To better understand these underlying classification failures, future work will extend our simulation experiments into real hardware environments. This transition will allow us to evaluate how real-time artifacts and distribution shifts impact measurement accuracy and classifier performance.

\section{Conclusion}

We introduced a closed-loop adaptive spectral measurement pipeline called BAARC, useful in scenarios with limited spectral coverage. It demonstrates that uncertainty-aware incremental measurement of spectral patches, along with stoppage conditioning, achieves classification performance close to that of full spectral acquisition while requiring only about 50\% of the total bandwidth. Our adaptive strategy significantly outperforms random-sampling baselines, validating the effectiveness of our uncertainty-driven scans for identifying discriminative chemical features. This helps with data measurement systems, e.g., real-world SRS scanning, where high-intensity lasers can cause sample degradation in live-cell imaging or with sensitive marine samples when long exposure times are required. It minimizes the laser exposure by stopping the laser as soon as the classifier is confident, preserving the sample for longitudinal studies. The availability of high-quality, classifiable training data is critical to our method, and therefore, we use four different benchmark Raman spectral data sets for demonstration. Future work will explore applying this framework to different optimization criteria across applications such as other spectroscopy systems~\citep{sparse_ap1, cs_raman}, atomic force microscopy~\citep{afm}, and mass spectrometry~\citep{ms} that demand adaptive sensing. By leveraging the inherent spectral redundancy of chemical manifolds, we can achieve "intelligent" spectrometers that optimize measurement effort based on the sample's complexity.


\section*{Code availability}
The code will be made available later after publishing.


\section*{Acknowledgement}
This study was funded by the European Union’s Horizon Europe research and innovation program for the project uCAIR with Grant Agreement No. 101135175.

\bibliographystyle{cas-model2-names}

\bibliography{references}



\clearpage
\appendix
\thispagestyle{empty}
\onecolumn
\begin{center}
    {\LARGE\bfseries Supplementary Material}
\end{center}

\vspace{1cm}

\section{Mutual Information as classification uncertainty} \label{ap:mi}
In~\autoref{fig:threshold} for threshold selection, we show how Entropy, as a measure of classification uncertainty, changes for the COVID-19 dataset. We can identify a similar change when using MI, as shown in~\autoref{fig:mi}, compared with~\autoref{fig:ent}. However, the threshold values are different.  
\begin{center}
    \includegraphics[width=0.5\linewidth]{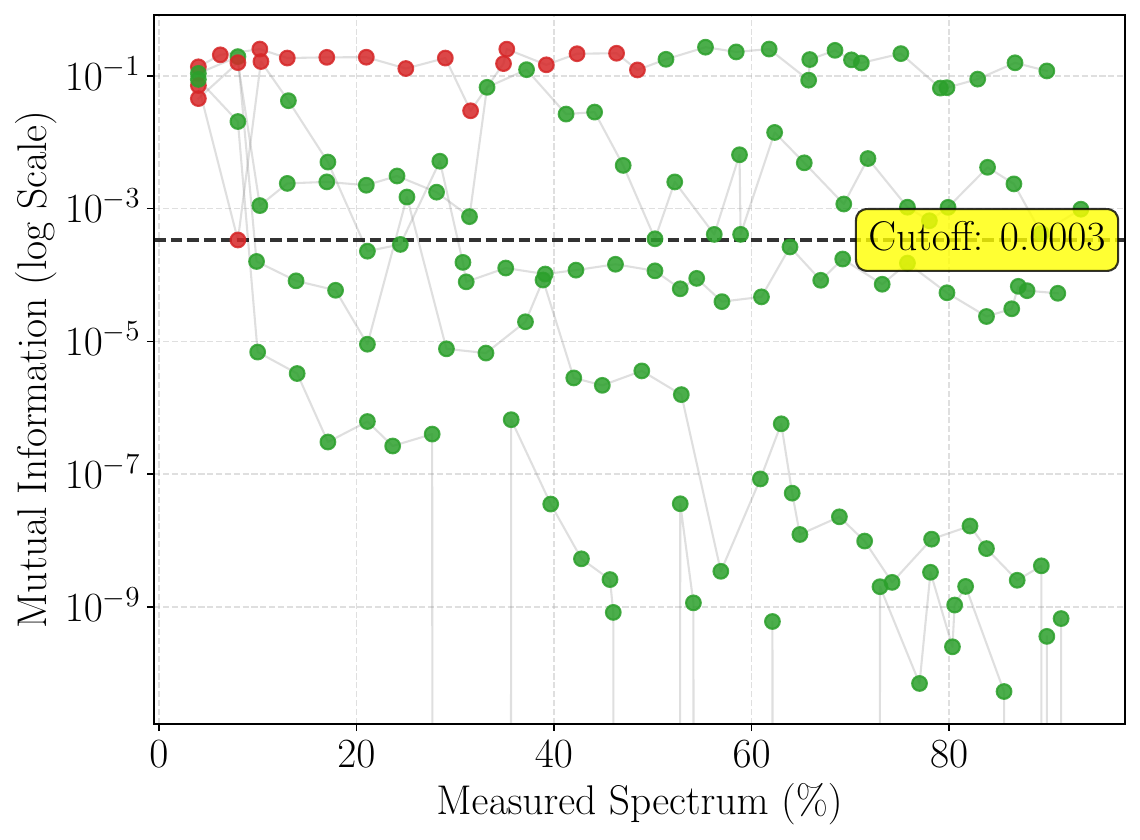}
    \captionof{figure}{Change in MI with percentage of measured spectrum for 5 COVID-19 spectra. This plot connects to the threshold selection process as discussed in~\autoref{sec:exp}.}
    \label{fig:mi}
\end{center}

\section{Dataset Summary} \label{app:data}
In~\autoref{tab:data} we show summary of dataset used in our experiments. 
\begin{center}
    \resizebox{\textwidth}{!}{%
    \begin{tabular}{c|cccc}
    \toprule
        Dataset & RRUFF & COVID-19 & Marine Pathology   &Bacteria\\
        \midrule
         Wavenumber range ($cm^{-1}$) & 76 - 1403  & 400 - 2112  & 600 - 1800   &382-1800\\
         No. of class & 1310 & 3 & 9  &30\\
         Dimentionality & 1328 & 900 & 1200  &1000\\
         Training/Tuning data & 7644 & 297  & 455  &54000/27000\\
         Testing data & 3554 & 93 & 570  &3000\\
         Application & Minerals Identification & Disease Diagnosis & Disease Diagnosis  &Disease Diagnosis\\
         \bottomrule
    \end{tabular}}
    \captionof{table}{Summary of datasets.}
    \label{tab:data}
\end{center}

\section{Hyperparameters and Thresholds} \label{ap:values}

The below~\autoref{tab:hyp} shows hyperparameters ($e$ = Epochs, $lr$ =  Learning rate) and threshold values ($\tau_{rec}$ = Reconstruction threshold
$\tau_{clf}$ = Classification threshold) for every dataset-model setting.

\begin{center}
    \resizebox{\textwidth}{!}{%
    \begin{tabular}{c|cccc}
    \toprule
        Dataset & RRUFF & COVID-19 & Marine Pathology   &Bacteria\\
        \midrule
         MCD & $e = 1000$, $lr = 10^{-4}$& $e = 1500$, $lr = 10^{-4}$&$e = 1000$, $lr = 10^{-4}$& $e = 700 + 500$, $lr = 10^{-4}$\\
          & $\tau_{rec} = 0.007$, $\tau_{clf}=0.1$& $\tau_{rec} = 0.003$, $\tau_{clf}=0.1$&$\tau_{rec} = 0.007$, $\tau_{clf}=0.1$&$\tau_{rec} = 0.007$, $\tau_{clf}=0.1$\\
          \midrule
          DGP & $e = 1000$, $lr = 10^{-3}$& $e = 1500$, $lr = 10^{-4}$&$e = 1000$, $lr = 10^{-4}$& $e = 700 + 500$, $lr = 10^{-4}$\\
          & $\tau_{rec} = 0.03$, $\tau_{clf}=0.1$& $\tau_{rec} = 0.03$, $\tau_{clf}=0.1$&$\tau_{rec} = 0.03$, $\tau_{clf}=0.1$&$\tau_{rec} = 0.007$, $\tau_{clf}=0.3$\\
         
          \midrule
         NP& $e = 1000$, $lr = 10^{-4}$& $e = 1500$, $lr = 10^{-4}$&$e = 1000$, $lr = 10^{-4}$& $e = 700 + 500$, $lr = 10^{-4}$\\
          & $\tau_{rec} = 0.03$, $\tau_{clf}=0.1$& $\tau_{rec} = 0.06$, $\tau_{clf}=0.1$&$\tau_{rec} = 0.03$, $\tau_{clf}=0.1$&$\tau_{rec} = 0.007$, $\tau_{clf}=0.1$\\
         \bottomrule
    \end{tabular}}
    \captionof{table}{All hyperparameter settings and threshold values for each dataset-model pair are listed, where changes from default values as discussed in the main paper can be noted.}
    \label{tab:hyp}
\end{center}

\section{Different models} \label{ap:sep_model}
Here, we show why combined training of reconstruction and classification models is necessary and show the advantage of adaptive sensing. To demonstrate this, we use a jointly trained TeaNet model and a separately trained model as in~\autoref{tab:sep}. We use a much more aggressive random fixed mask of 80\% on the spectra to demonstrate the drawback. We clearly see that, even though the reconstruction performance of the jointly trained model is lower due to its joint optimization criterion, its classification performance is consistent. The main insight is obtained from the classification performance on the COVID-19 dataset. Unlike other datasets, this dataset shows fine changes in specific regions of the spectra~\citep{data_cov}. With the masking, the model's predictions always converge to a single class. This overfitting behavior of the reconstruction model affects the classification done on it. This is evident from COVID-19 results. It has 3 classes in total, with test set class distribution of 30, 31, and 32 for classes 0 (healthy), 1 (suspected), and 2 (COVID-19), respectively. We see that class 1 is always predicted, resulting in 33.33\% accuracy across all runs. To avoid this, we use joint training. In adaptive sensing, this behavior is rarely observed because the sensing methodology prioritizes measuring such contrasting regions.

\begin{center}

    \begin{tabular}{c|ccc}
    \toprule
        Models & RRUFF & COVID-19 & Marine Pathology   \\
        \midrule
         Joint Rec MSE& 0.02438 ± 0.0004& 0.01044 ± 1.5163& 0.02890 ± 0.0004 \\
         Separate Rec MSE & 0.00309 ± 0.0008& 0.00160 ± 0.0009&0.00733 ± 0.0004 \\
         \midrule
         Joint Clf Accuracy & 71.04\% ± 0.73
        & 64.16\% ± 2.82&81.93\% ± 0.38 \\
                 Separate Clf Accuracy & 87.39\% ± 3.74
        & 33.33\% ± 0
        & 82.28\% ± 0.34 \\
         \bottomrule
    \end{tabular}
    \captionof{table}{Comparison of joint training vs separately trained model.}
    \label{tab:sep}
\end{center}

\section{Reconstruction performance} \label{ap:rec_app}

In~\autoref{tab:res_rec}, we quantitatively compare the reconstruction of adaptive and random models across various models, using MSE for error and correlation for similarity between the reconstruction and the ground truth. We see that adaptive and random perform similarly across most models, which contradicts the qualitative analysis focusing on fine-grained classification. This is also evident from our classification analysis. However, from the~\autoref{tab:res_rec} we can gather that NP has the highest performance over all datasets and DGP has the least reconstruction performance. 

\renewcommand{\arraystretch}{1.2}
\begin{center}
\centering
\resizebox{\textwidth}{!}{%
\begin{tabular}{ll|cc|cc|cc|cc}
\toprule
 & Dataset: & \multicolumn{2}{c|}{\textbf{Marine pathology}} & \multicolumn{2}{c|}{\textbf{COVID-19}} & \multicolumn{2}{c|}{\textbf{RRUFF}} & \multicolumn{2}{c}{\textbf{Bacteria}}\\
\midrule
Model & Measurement: & Adaptive & Random & Adaptive & Random & Adaptive & Random & Adaptive & Random \\
\midrule
\multirow{2}{*}{\textbf{MCD}} & MSE & $(5.273 \pm 2.696)\times 10^{-3}$ & $(6.90 \pm 3.45)\times 10^{-3}$ & $(8.50 \pm 0.36)\times 10^{-4}$ & $(4.917 \pm 0.125)\times 10^{-3}$ & $(9.22 \pm 0.015)\times 10^{-3}$ & $(8.04 \pm 0.040)\times 10^{-3}$ & $(4.44 \pm 2.22)\times 10^{-3}$ & $(4.77 \pm 2.40)\times 10^{-3}$ \\
 & Corr & 93.85$\pm$0.08 & 90.70$\pm$0.13 & 97.67$\pm$0.08 & 97.85$\pm$0.01 & 56.02$\pm$0.68 & 63.78$\pm$0.13 & 89.33$\pm$0.47 & 90.0$\pm$0.81 \\
\midrule
\multirow{2}{*}{\textbf{DGP}} & MSE & $(5.22 \pm 0.07)\times 10^{-2}$ & $(4.93 \pm 0.09)\times 10^{-2}$ & $(1.59 \pm 0.19)\times 10^{-2}$ & $(1.48 \pm 0.27)\times 10^{-2}$ & $(5.27 \pm 0.08)\times 10^{-3}$ & $(4.94 \pm 0.03)\times 10^{-3}$ & $(9.00 \pm 4.51)\times 10^{-2}$ & $(8.75 \pm 4.38)\times 10^{-2}$ \\
 & Corr & 48.03$\pm$0.32 & 43.8$\pm$9.2 & 60.3$\pm$4.73 & 62.21$\pm$6.78 & 82.72$\pm$0.34 & 80.67$\pm$0.23 & 47.80$\pm$2.12 & 37.66$\pm$0.94 \\

\midrule
\multirow{2}{*}{\textbf{NP}} & MSE & $(2.14 \pm 0.05)\times 10^{-4}$ & $(2.98 \pm 0.18)\times 10^{-4}$ & $(3.10 \pm 0.16)\times 10^{-6}$ & $(1.81 \pm 0.07)\times 10^{-5}$ & $(5.90 \pm 2.60)\times 10^{-4}$ & $(2.55 \pm 1.27)\times 10^{-3}$ & $(6.16 \pm 3.09)\times 10^{-4}$ & $(1.90 \pm 1.15)\times 10^{-4}$ \\
 & Corr & 94.36$\pm$0.06 & 92.74$\pm$0.27 & 99.62$\pm$0.01 & 99.17$\pm$0.04 & 97.31$\pm$0.05 & 89.32$\pm$0.33 & 98.56$\pm$0.09 & 99.63$\pm$0.09 \\
\bottomrule
\end{tabular}}
\captionof{table}{Reconstruction performance of adaptive and random sensing pipeline for four datasets: Marine pathology~\citep{data_marine}, COVID-19~\citep{data_cov}, RRUFF~\citep{data_rruff}, and Bacteria~\citep{bacteria} with models MC Dropout, Neural Process, and Deep Gaussian Process. We compare the MSE and the correlation between the intermediate reconstructed spectra and the complete actual spectra.}
\label{tab:res_rec}
\end{center}

\section{Reconstruction steps} \label{ap:rec}
\autoref{fig:s1} to~\autoref{fig:s13} show how reconstruction changes with measurement steps for the Marine Pathology dataset. Here, we show some intermediate steps before making a final decision, along with the reconstruction uncertainty as variance across the spectra. The region with the highest variance observed in one step will be measured next. The model barely predicts the correct peaks in step one, and this changes by the end of seven steps. Later, we refine our reconstructions with additional measurement steps. We can also see the uncertainty values changing at each step. One major finding from this example is that at step seven, our model already predicts class 0, the true class. However, we see that the model is confused at step 11, which is later corrected in step 12 as the uncertainty (Entropy) drops significantly.

\begin{center}
    \centering
    \includegraphics[width=0.9\textwidth]{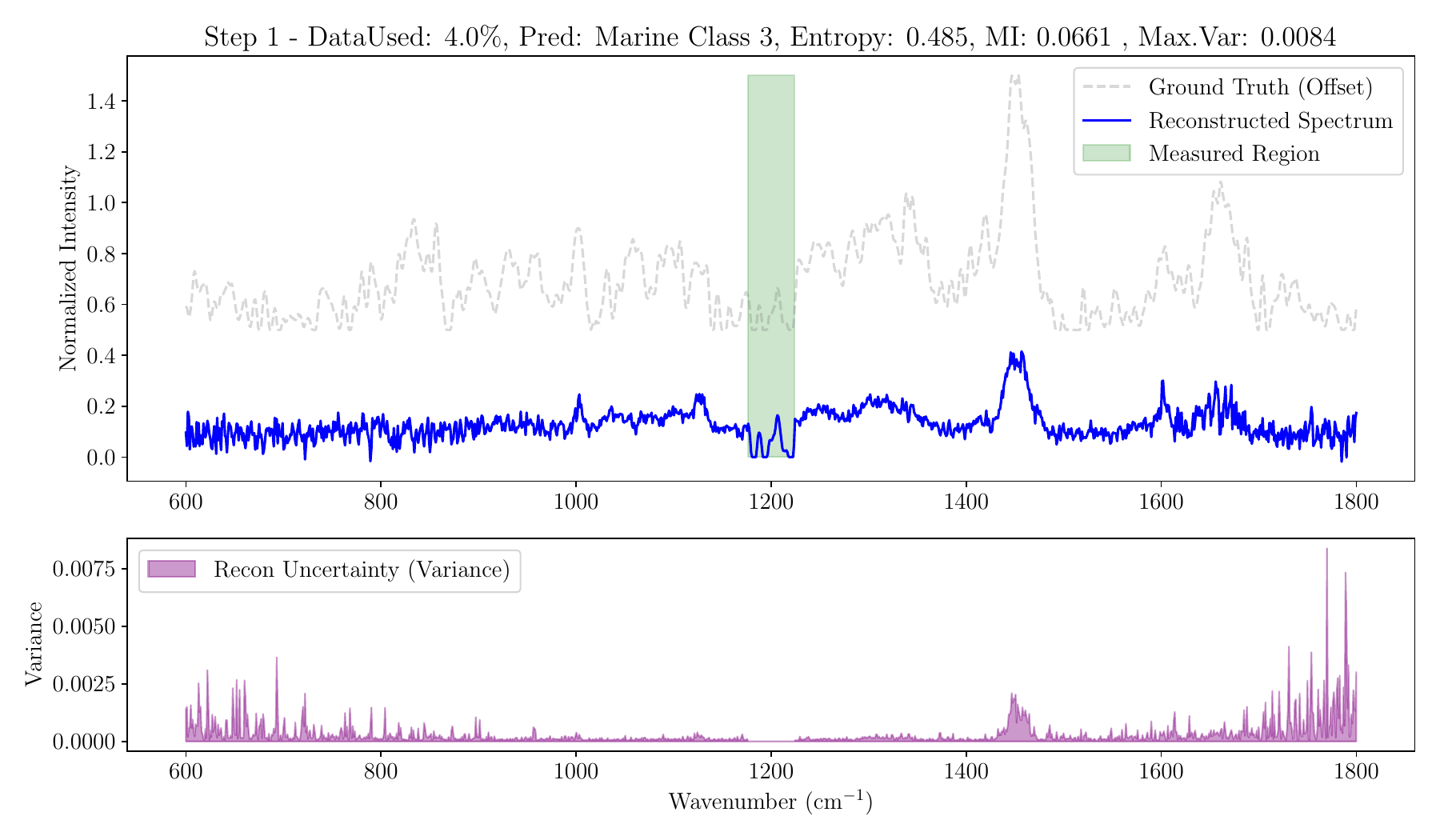}
    \captionof{figure}{Measurement step 1}
    \label{fig:s1}
\end{center}

\begin{center}
    \centering
    \includegraphics[width=0.9\textwidth]{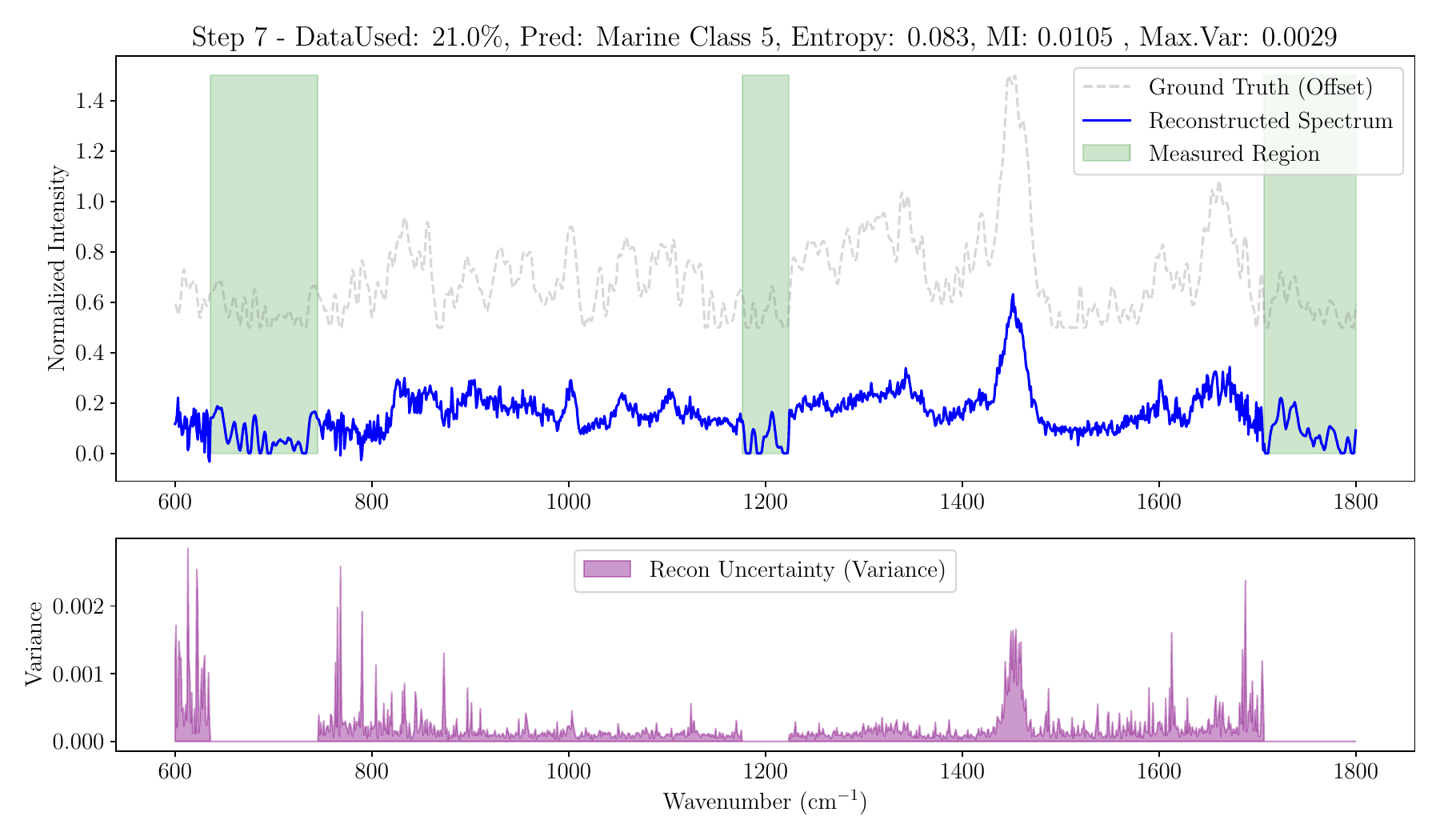}
    \captionof{figure}{Measurement step 7}
    \label{fig:s7}
\end{center}

\begin{center}
    \centering
    \includegraphics[width=0.9\textwidth]{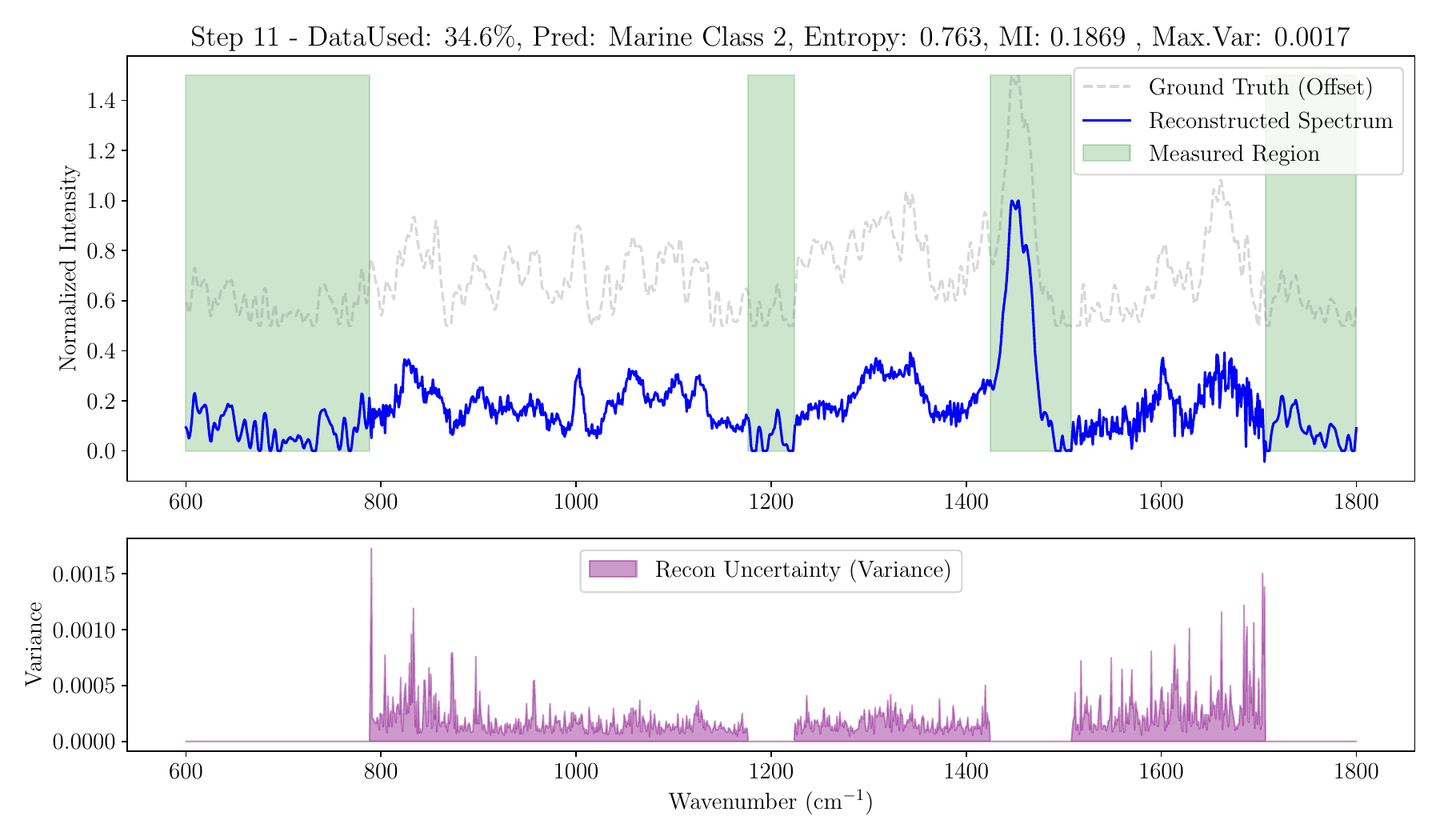}
    \captionof{figure}{Measurement step 11}
    \label{fig:s12}
\end{center}

\begin{center}
    \centering
    \includegraphics[width=0.9\textwidth]{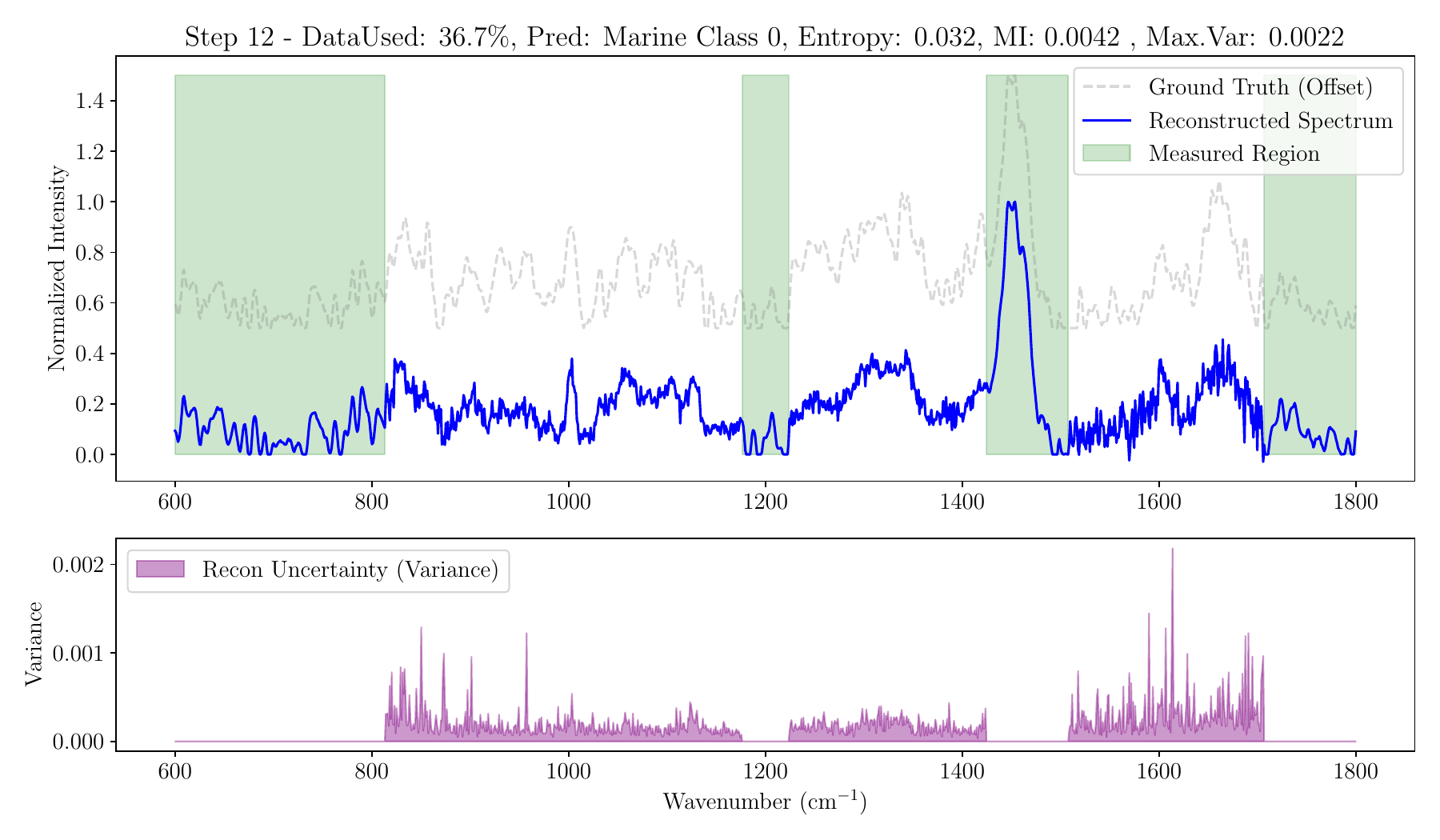}
    \captionof{figure}{Measurement step 12}
    \label{fig:s13}
\end{center}

\end{document}